\documentclass[10pt,twocolumn,letterpaper]{article}

\usepackage{cvpr}              

\usepackage{times}
\usepackage{epsfig}
\usepackage{graphicx}
\usepackage{amsmath}
\usepackage{amssymb}
\usepackage{csquotes}

\usepackage{multirow}
\usepackage{graphicx}
\usepackage{booktabs}
\usepackage{bbm}

\usepackage{color, colortbl}
\newcommand*{\twoelementtable}[3][l]%
{%
\renewcommand{\arraystretch}{0.8}%
\begin{tabular}[t]{@{}#1@{}}%
#2\tabularnewline
#3%
\end{tabular}%
}
\usepackage{array}
\usepackage{tabularx}
\usepackage[usenames,dvipsnames]{xcolor}
\newcolumntype{C}[1]{>{\centering\arraybackslash}p{#1}}
\newcommand{\RN}[1]{%
  \textup{\uppercase\expandafter{\romannumeral#1}}%
}

\usepackage[pagebackref=true,breaklinks=true,colorlinks,bookmarks=false]{hyperref}

\usepackage[capitalize]{cleveref}
\crefname{section}{Sec.}{Secs.}
\Crefname{section}{Section}{Sections}
\Crefname{table}{Table}{Tables}
\crefname{table}{Tab.}{Tabs.}

\def\cvprPaperID{621} 
\def\confName{CVPR}
\def\confYear{2022}

\begin{document}

\title{Unbiased Subclass Regularization for Semi-Supervised Semantic Segmentation}

\author{Dayan Guan, Jiaxing Huang, Aoran Xiao, Shijian Lu\thanks{Corresponding author.} \\ 
Singtel Cognitive and Artificial Intelligence Lab for Enterprises, Nanyang Technological University\\
{\tt\small \{Dayan.Guan, Jiaxing.Huang, Aoran.Xiao, Shijian.Lu\}@ntu.edu.sg}
}

\maketitle

\begin{abstract}
Semi-supervised semantic segmentation learns from small amounts of labelled images and large amounts of unlabelled images, which has witnessed impressive progress with the recent advance of deep neural networks. However, it often suffers from severe class-bias problem while exploring the unlabelled images, largely due to the clear pixel-wise class imbalance in the labelled images. This paper presents an unbiased subclass regularization network (USRN) that alleviates the class imbalance issue by learning class-unbiased segmentation from balanced subclass distributions. We build the balanced subclass distributions by clustering pixels of each original class into multiple subclasses of similar sizes, which provide class-balanced pseudo supervision to regularize the class-biased segmentation. In addition, we design an entropy-based gate mechanism to coordinate learning between the original classes and the clustered subclasses which facilitates subclass regularization effectively by suppressing unconfident subclass predictions. Extensive experiments over multiple public benchmarks show that USRN achieves superior performance as compared with the state-of-the-art. 
\end{abstract}

\section{Introduction}

Semantic segmentation aims to assign a human-defined class label to each pixel of an image, which is a fundamental task in the computer vision research. With the recent advance of deep neural networks~\cite{he2016deep,zhao2017pyramid,chen2018encoder}, we can learn a very accurate segmentation model while a large amount of labelled training images are available. However, collecting a large amount of pixel-wise semantic labels is laborious and time-consuming, which has become a bottleneck in semantic segmentation research~\cite{everingham2010pascal,lin2014microsoft,cordts2016cityscapes}. Semi-supervised semantic segmentation, which aims to learn from a small amount of labelled images and a large amount of unlabelled images, has been attracting increasing attention for addressing the image annotation challenge.
\begin{figure}[t]
\centering
\includegraphics[width=1.0\linewidth]{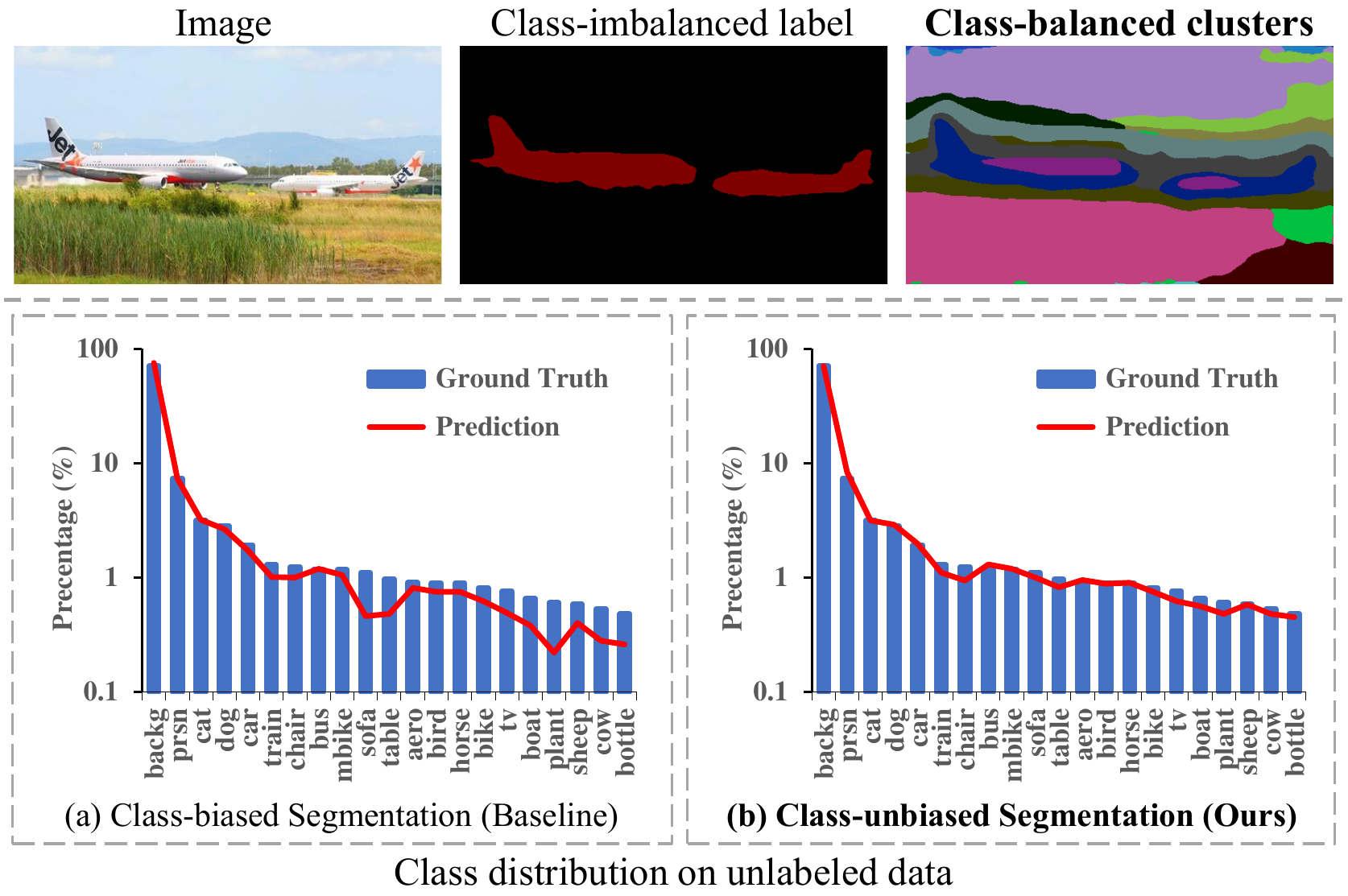}
\caption{
Motivation of our work: In semi-supervised semantic segmentation, the segmentation model trained using \textit{class-imbalanced labels} (of the labelled data) tends to produce \textit{class-biased segmentation} on the unlabelled data. We create \textit{class-balanced clusters} with balanced subclass distribution, learning from which alleviates the class imbalance issue and produces \textit{class-unbiased segmentation} on the unlabelled data. We obtain the class-balanced clusters by clustering pixels of each original class into multiple subclasses of similar size. Best viewed in color. 
}
\label{fig:intro}
\end{figure}

Most existing studies tackle the challenge of semi-supervised semantic segmentation by applying either consistency-training~\cite{ouali2020semi,mittal2019semi,lai2021semi} or self-training~\cite{mendel2020semi, luo2020semi,chen2021semi,huo2021atso,Yuan_2021_ICCV,he2021re} to the unlabelled data. However, they often suffer from constrained segmentation accuracy largely due to the segmentation model that is trained by using the labelled data. As illustrated in Fig.~\ref{fig:intro}, the model trained using the labelled data is class biased due to the class-imbalance of the labelled data. This leads to class-biased segmentation of the unlabelled data which accumulates and finally degrades the whole semi-supervised learning. Though a few studies~\cite{he2021re,wei2021crest} attempt to handle the class imbalance issue by selecting more pseudo labels for minority classes during self-training, these pseudo labels are often noisy as they are generated from class-biased segmentation. Note the class imbalance issue has been widely studied in supervised learning via re-sampling~\cite{chawla2002smote, buda2018systematic, wang2019dynamic, kim2020m2m, sohoni2020no}, re-weighting~\cite{huang2016learning, lin2017focal, ren2018learning, cui2019class} and meta-learning~\cite{wang2017learning, ren2018learning, shu2019meta}, but these works require labels to rectify biased predictions and are thus inapplicable to the unlabelled data in semi-supervised semantic segmentation.

In this work, we propose an unbiased subclass regularization network (USRN) that tackles the class-imbalance issue and regularizes class-biased segmentation by \textit{generating} class-unbiased segmentation. Leveraging the segmentation backbone as learnt from the original class distribution, USRN introduces an auxiliary segmentation task as supervised by a set of class-balanced clusters for producing class-unbiased segmentation on the unlabelled data. We obtain the class-balanced clusters from the labelled data by clustering pixels of each original class into multiple subclasses of similar size. As illustrated in Fig.~\ref{fig:intro}, the USRN trained using class-balanced clusters can produce clearly more class-unbiased segmentation for the unlabelled data. In addition, the segmentation with the original classes could be interfered by the segmentation with the generated subclasses due to their different convergence speeds. We design an entropy-based gate mechanism to address this issue, where the learning with the auxiliary subclasses will be stopped (i.e., no back-propagation) when the subclass predictions are less confident than the original class predictions. Extensive experiments over multiple public benchmarks demonstrate the effectiveness of our designed network.

The contribution of this work is threefold. \textit{First}, we propose an unbiased subclass regularization network that explores class-unbiased segmentation to alleviate the class imbalance issue in semi-supervised semantic segmentation. \textit{Second}, we design an entropy-based gate mechanism that coordinates the concurrent learning from the original classes and the generated subclasses effectively. \textit{Third}, extensive experiments show the superior effectiveness of our designed network as compared with the state-of-the-art.

\begin{figure*}[!t]
\centering
\includegraphics[width=1.0\linewidth]{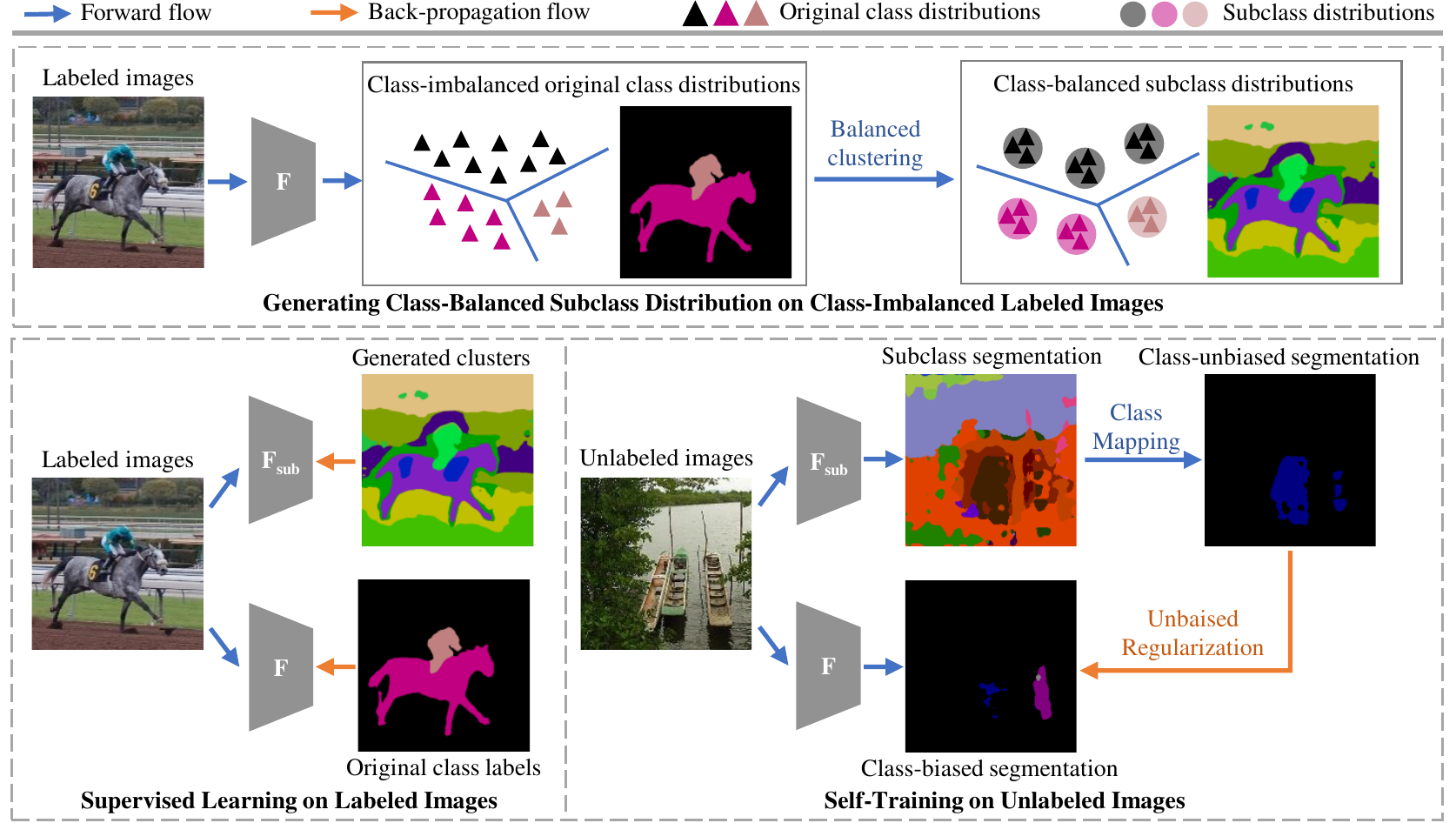}
\caption{
Overview of unbiased subclass regularization network (USRN): USRN regularizes \textit{class-biased segmentation} from original class distribution with \textit{class-unbiased segmentation} from subclass distributions. We generate \textit{class-balanced subclass distribution} by clustering features from \textit{class-imbalanced original class distribution} into multiple groups of similar size. Specifically, USRN performs supervised learning by training a class-biased model $F$ and a class-unbiased model $F_{sub}$ with \textit{original class labels} and \textit{generated clusters}, respectively, for labelled images under semi-supervised setup. 
For unlabelled images, USRN performs self-training by applying \textit{class-unbiased segmentation} from $F_{sub}$ to regularize \textit{class-biased segmentation} from $F$.
We obtain class-unbiased segmentation by mapping the \textit{subclass segmentation} (as produced by  $F_{sub}$) from subclass space to original class space. Best viewed in color.
}
\label{fig:method}
\end{figure*} 

\section{Related Works}

\subsection{Supervised Semantic Segmentation}
With recent progress of deep learning, supervised semantic segmentation has made remarkable progress by designing various  architectures. FCN in~\cite{long2015fully} is the first end-to-end trainable network with fully convolutional layers for semantic segmentation. The subsequent studies improve \cite{long2015fully} by employing encoder-decoder structures \cite{badrinarayanan2017segnet,chen2018encoder,ronneberger2015u}, mutli-scale inputs~\cite{chen2016attention,dai2015convolutional,lin2016efficient}, feature pyramid spatial pooling~\cite{liu2015parsenet,zhao2017pyramid}, attention mechanism~\cite{zhao2018psanet,fu2019dual} or dilated convolutions~\cite{yu2015multi,chen2017deeplab,wang2018understanding,chen2018encoder}. For example, Deeplabv3+ in~\cite{chen2018encoder} combines low-level and high-level features to refine the object boundaries of segmentation results. However, training these supervised segmentation networks requires large amounts of annotated data which is often laborious and time-consuming to collect. Our work aims to alleviate the data annotation constraint by exploring large amounts of unlabeled data together with limited amount of labeled data.

\subsection{Semi-Supervised Semantic Segmentation}
Semi-supervised segmentation aims to explore tremendous unlabeled data with supervision from limited labeled data, which is most relevant to domain adaptive segmentation where labeled data is obtained from another domain~\cite{vu2019advent,huang2020contextual,yang2020fda,huang2021rda,guan2021domain,araslanov2021self,huang2022multi}. Most existing studies address this challenge by either consistency-training~\cite{qi2019ke,zhou2021c3,zhong2021pixel,tarvainen2017mean,ke2019dual,guan2021scale,xie2020unsupervised,huang2021model,li2021comatch} or self-training~\cite{ibrahim2020semi,zou2021pseudoseg,berthelot2019mixmatch,sohn2020fixmatch,kim2020distribution,ren2020not,huo2021atso,huang2021fsdr,Alonso_2021_ICCV,huang2021cross,taherkhani2021self,hu2021semi}. Specifically, consistency-training maintains the consistency of segmentation of each unlabeled sample under different perturbations. For example, CCT~\cite{ouali2020semi} applies two same-structured segmentation networks with different initialization to produce differently perturbed samples. CAC~\cite{lai2021semi} enforces context-aware consistency between representations from the same unlabeled image augmented with different context information. Self-training instead generates pseudo labels on unlabeled data to re-train networks. For example, GCT~\cite{ke2020guided} introduces a flaw detector to correct the defects in pseudo labels.  DBSN~\cite{Yuan_2021_ICCV} designs distribution-specific batch normalization for robust pseudo labels generation. CPS~\cite{chen2021semi} produces pseudo labels from one segmentation network to supervise the other segmentation network with the same structure yet different initialization. However, both consistency-training and self-training suffer from clear pixel-wise class imbalance in labeled data. Our method can mitigate the class imbalance issue in semi-supervised segmentation effectively.

\subsection{Class-Imbalance Learning}
Class imbalance issue has been widely studied in supervised learning. For example, re-sampling based methods~\cite{chawla2002smote,buda2018systematic,wang2019dynamic,kim2020m2m} re-balance the biased networks according to sample sizes for each class. Re-weighting based methods~\cite{huang2016learning,lin2017focal,ren2018learning,cui2019class} adaptively adjust the loss weight for different training samples with different class labels. Meta-learning based methods~\cite{wang2017learning,ren2018learning,shu2019meta} use the validation loss calculated from selected class-balanced labeled samples as the meta objective to optimize networks. However, all these methods rely on labels to address the class imbalance issue and cannot be directly applied to unlabeled data in semi-supervised learning. Recently, several studies attempt to handle the class imbalance issue in semi-supervised learning. For example, CReST~\cite{wei2021crest} selects pseudo labels more frequently for minority classes according to the estimated class distribution. DARS~\cite{he2021re} employs adaptive threshold to select more pseudo labels for minority class during self-training.
However, these methods tends to generate noisy pseudo labels from class-biased segmentation of unlabeled data. We address the class imbalance issue by constructing and learning from class-balanced subclasses.

\section{Method}

\subsection{Problem Definition}
This work focuses on semi-supervised semantic segmentation. Given images $X_{l} \subset \mathbb{R}^{H \times W \times 3}$ with pixel-level semantic labels $\hat{y} \subset (1,C)^{H \times W}$ and unlabelled images $X_{u} \subset \mathbb{R}^{H \times W \times 3}$ ($H$, $W$ and $C$ denote image height, image width and class number, respectively), the goal is to learn a segmentation model $F$ that can fit both labelled and unlabelled data and work well on unseen images. Existing methods~\cite{ouali2020semi,ke2020guided,lai2021semi,mendel2020semi, luo2020semi, chen2021semi,huo2021atso,zou2021pseudoseg,he2021re} combine supervised learning on labelled images and unsupervised learning on unlabelled image to tackle the semi-supervised challenge. For labelled images, they adopt cross entropy loss as supervised loss $\mathcal{L}_{s}$ to train $F$. For unlabelled images, they adopt consistency regularization loss~\cite{ouali2020semi,ke2020guided} or self-training loss~\cite{mendel2020semi,luo2020semi,lai2021semi,chen2021semi,huo2021atso, zou2021pseudoseg,he2021re} as unsupervised loss $\mathcal{L}_{u}$ to train $F$. The overall objective is a weighted combination of supervised and unsupervised losses:
\begin{equation}
\mathcal{L} = \mathcal{L}_{s} (X_{l},Y)+\lambda_{u} \mathcal{L}_{u} (X_{u}),
\label{eq:obj}
\end{equation}
where $\lambda_{u}$ is a balancing weight. With this objective function, supervised and unsupervised learning would benefit each other thanks to their complementary nature~\cite{van2020survey}.

Though consistency-training and self-training can learn from the unlabelled images effectively, their performance is often constrained by the quality of the supervised model that is trained by using the labelled images. Specifically, the labelled images often suffer from a clear class-imbalance issue which directly leads to class-biased model and further class-biased segmentation on the unlabelled images. Such class-biased segmentation accumulates during the process of consistency-training or self-training which finally degrades the overall performance of semi-supervised semantic segmentation. We define this problem as a class-imbalance issue in semi-supervised semantic segmentation, and design a class-balanced subclass regularization network to address the class-imbalance problem.

\subsection{Unbiased Subclass Regularization}
We design an unbiased subclass regularization network (USRN) for addressing the class-imbalance issue in semi-supervised segmentation, as shown in Fig.~\ref{fig:method}. With labelled images in semi-supervised segmentation, USRN first trains a class-biased model $F$ (by learning from the class-imbalanced labelled images) and then produces a class-balanced subclass distribution by clustering the F-produced features of the labelled images. With the class-balanced subclass distribution, a class-unbiased model $F_{sub}$ can be trained which tends to produce class-unbiased segmentation while applied to unlabelled images in semi-supervised segmentation. 

\noindent\textbf{Generating class-balanced subclass distribution.} 
USRN learns class-unbiased model by generating class-balanced clusters. With the labelled images (with class imbalanced annotations), USRN first trains a supervised segmentation model $F$ and then applies $F$ to each labelled image to extract semantic features. It then adopts balanced k-means clustering~\cite{lin2019balanced} to group the extracted semantic features into multiple clusters of similar size. The generated class-balanced clusters $\hat{y}^\star \subset (1,C_{sub})^{H \times W}$ ($C_{sub}$ is the number of clustered subclasses) directly give a balanced subclass distribution with the labelled images. In our implementation, we empirically set the cluster size as the size of the smallest class in the original annotations.

\noindent\textbf{Supervised learning on labeled data.}  
USRN performs supervised learning with both original and subclass annotations. For each labelled image $x_l$, we feed a weakly augmented image $\mathcal{A}^{w}(x_{l})$ to $F$ to obtain original class prediction $p_{l}^{w} = F(\mathcal{A}^{w}(x_{l}))$ and the same input to $F_{sub}$ to obtain subclass prediction $p_{l}^{w\star} = F_{sub}(\mathcal{A}^{w}(x_{l}))$. Here, $\mathcal{A}^{w}$ is a weak augmentation function, \ie, random scaling, cropping and horizontal flipping. Given $p_{l}^{w}$ with its original class label $\hat{y} \subset Y$ and $p_{l}^{w\star}$ with its class-balanced cluster $\hat{y}^\star \subset Y^\star$, a multi-distribution supervised loss $\mathcal{L}_{s}^{md}$ can be defined by: 
\begin{equation}
\mathcal{L}_{s}^{md} = \mathcal{L}_{ce} (p_{l}^{w},\hat{y})+ \lambda_{sub} \mathcal{L}_{ce} (p_{l}^{w\star},\hat{y}^\star),
\label{loss:md}
\end{equation}
where $\mathcal{L}_{ce}$ is cross-entropy loss and $\lambda_{sub}$ is a balancing weight.

\noindent\textbf{Self-training on unlabeled data.} USRN preforms self-training to update $F$ with class-unbiased pseudo label as generated from the subclass distribution. For each unlabelled sample $x_u$, we feed a weakly augmented image $\mathcal{A}^{w}(x_{u})$ to $F$ to obtain original class prediction $p_{u}^{w} = F(\mathcal{A}^{w}(x_{u}))$ and the same input to  $F_{sub}$ to obtain subclass prediction $p_{u}^{w\star} = F_{sub}(\mathcal{A}^{w}(x_{u}))$. To generate unbiased pseudo labels for the original class supervision, we first map the prediction $p_{u}^{w\star}$ from the subclass space $(1,C_{sub})^{H \times W}$ to the original class space $(1,C)^{H \times W}$ (this process denoted by $\mathcal{M}$), and then define a function $\mathcal{S}$ to select pseudo labels from the mapped predictions in an online manner. We define the pseudo-label selection function $\mathcal{S}$ by:
\begin{equation}
\begin{split}
\mathcal{S}(p) = \mathbbm{1}_{[p^{(c)} > \gamma]}(p^{(c)}),
\end{split}
\label{eq:s}
\end{equation}
where $p$ refers to the predictions, $\mathbbm{1}$ is a function that returns the class index $c$ if the condition is true or the `ignore' class index otherwise, and $\gamma$ is a confidence threshold. Note there is no back-propagation for the `ignore' class in training.

To alleviate over-fitting in self-training, the pseudo label generated from weakly augmented version of an image $\mathcal{A}^{w}(x_{u})$ is used to supervise the segmentation from the strongly augmented version of the same image $\mathcal{A}^{s}(x_{u})$. Here, $\mathcal{A}^{s}$ is a strong augmentation function, \ie, random color jitters and Gaussian blur. 
With $p_{u}^{w}$ and $\hat{p}_{u}^{w\star}$ (one-hot vector computed from $p_{u}^{w\star}$ using softmax) from $\mathcal{A}^{w}(x_{u})$, 
we simultaneously feed $\mathcal{A}^{s}(x_{u})$ to $F$ to obtain the original class prediction $p_{u}^{s} = F(\mathcal{A}^{s}(x_{u}))$, and perform subclass regularized self-training with the loss $\mathcal{L}_{st}$:
\begin{equation}
\mathcal{L}_{st} = \mathcal{L}_{ce} (p_{u}^{s},\mathcal{S}(\mathcal{M}(\hat{p}_{u}^{w\star})\cdot{}p_{u}^{w}))
\label{loss:st}
\end{equation}

In addition, USRN performs self-training on subclass distributions to update $F_{sub}$. 
With $p_{u}^{w\star}$ from $p_{u}^{w\star}$ as in Eq.~\ref{loss:st}, 
we simultaneously feed $\mathcal{A}^{s}(x_{u})$ to $F_{sub}$ to obtain the subclass prediction $p_{u}^{s\star}$, and perform subclass self-training with the loss $\mathcal{L}_{st}$:
\begin{equation}
\mathcal{L}_{st}^{sub} = \mathcal{L}_{ce} (p_{u}^{s\star},\mathcal{S}(p_{u}^{w\star})).
\label{loss:sub_st}
\end{equation}

\subsection{Entropy-based Gate Mechanism}
The proposed USRN employs subclass predictions to regularize original-class predictions. As the subclass distributions are derived from the original-class distribution, learning from the subclass distributions is more complex and tends to be slower as compared with that from the original-class distributions under the same learning policy (\eg, optimizer, learning rate, weight decay rate, etc). This could introduce undesired regularization. Specifically, the original-class learning could produce more confident and correct predictions than the subclass learning as the original-class learning converges faster than the subclass learning in training. The semi-supervised learning will degrade if the original-class predictions are regularized by the subclass predictions under such circumstance.

To address this problem, we design an entropy-based selection function to avoid regularizing the confident original-class predictions $p$ with unconfident subclass predictions $p^{\star}$. The entropy-based selection function is defined by:
\begin{equation}
\mathcal{S}_{e}(p^{\star},p) = \mathbbm{1}_{[\mathcal{E}(p^{\star}) < \mathcal{E}(p)]}(\mathcal{S}(\hat{p}^{\star})\cdot{}p),
\end{equation}
where $\mathcal{E}$ is the entropy function as defined in~\cite{shannon1948mathematical}.

Given the original predictions (\ie, $p_{u}^{s}$ and $p_{u}^{w}$ from strongly and weakly augmented versions of the same image) and the subclass prediction (\ie, $p_{u}^{w\star}$ from the weakly augmented version) as in Eq.~\ref{loss:st}, we reformulate the self-training loss in Eq.~\ref{loss:st} and define an entropy-based self-training loss $\mathcal{L}_{st}^{e}$ as follows:
\begin{equation}
\mathcal{L}_{st}^{e} = \mathcal{L}_{ce} (p_{u}^{s}, \mathcal{S}_{e}(\mathcal{M}(p_{u}^{w\star}),p_{u}^{w}))
\label{loss:st_eg}
\end{equation}

Combining the losses in Eqs.~\ref{loss:md},~\ref{loss:sub_st} and~\ref{loss:st_eg}, the overall training objective of the unbiased subclass regularization network (USRN) can be formulated as follows:
\begin{equation}
\mathcal{L}_{\text{USRN}} = \mathcal{L}_{s}^{md}+\lambda_{u} (\mathcal{L}_{st}^{e}+\lambda_{sub} \mathcal{L}_{st}^{sub}).
\end{equation}

\section{Experiments}

\subsection{Experimental Setting}
\noindent\textbf{Datasets.}
We conducted main experiments on the dataset PASCAL VOC~\cite{everingham2010pascal} by following previous work~\cite{ke2020guided,ouali2020semi,lai2021semi,chen2021semi}. The dataset consists of 10,582 images for training and 1,456 images for evaluation, and the image resolution varies from $192\times{}282$ to $500\times{}500$. It provides pixel-wise annotations with 21 semantic classes. To perform comprehensive validation, we also conducted experiments on the dataset Cityscapes~\cite{cordts2016cityscapes} which contains 2,975 images for training and 500 images for evaluation and all images have the same resolution of $1024\times2048$. Cityscapes provides pixel-wise labels with 19 semantic classes.

\noindent\textbf{Implementation details.}
Both segmentation backbone model $F$ and auxiliary segmentation model $F_{sub}$ adopt Deeplabv3+~\cite{chen2018encoder} with ResNet-50~\cite{he2016deep} pre-trained on ImageNet~\cite{deng2009imagenet}, where $F$ and $F_{sub}$ share layers that extract low-level features in ResNet-50. All network models are optimized by mini-batch stochastic gradient descent (SGD) with a base learning rate of $10^{-3}$, a momentum of 0.9 and a weight decay of $10^{-4}$. The weak augmentation function $\mathcal{A}^{w}$ (\ie, random scaling, cropping and horizontal flipping) and the strong augmentation function $\mathcal{A}^{s}$ (\ie, random color jitters and Gaussian blur) are the same as in~\cite{lai2021semi}. The confidence threshold $\gamma$ is set to 0.75 and all the balancing weights (\ie, $\lambda_{sub}$ and $\lambda_{u}$) are directly set to 1. During evaluation, each image is tested only on the segmentation backbone and the mean intersection-over-union (mIoU) is adopted as the evaluation metric.

\renewcommand\arraystretch{1.16}
\begin{table}[!ht]
\centering
\begin{footnotesize}
\begin{tabular}{p{2.0cm}|C{1.2cm}|*{3}{C{0.8cm}}}
\toprule
Method & Publication  &1/64 &1/32 &1/16 \\
\midrule
Baseline & -  &52.4  &59.2  &63.9  \\
GCT~\cite{ke2020guided} & ECCV 20  &-  &- &64.1  \\
CCT~\cite{ouali2020semi} & CVPR 20  &-  &-  &65.2  \\
DARS~\cite{he2021re}  & ICCV 21  &56.9  &64.5  &68.4  \\
DBSN~\cite{Yuan_2021_ICCV} & ICCV 21  &57.5  &64.6  &69.8  \\
CAC~\cite{lai2021semi} & CVPR 21  &56.5  &65.1  &70.1  \\
CPS~\cite{chen2021semi} & CVPR 21  &57.9  &64.8  &68.2  \\
\textbf{USRN (Ours)} & -  &\textbf{61.7}  &\textbf{68.6}  &\textbf{72.3}  \\ 
\hline Oracle & -  &76.8  &76.8  &76.8  \\
\bottomrule
\end{tabular}
\end{footnotesize}
\caption{Quantitative comparison with the state-of-the-art over the dataset PASCAL VOC~\cite{everingham2010pascal}. We randomly split 1/64, 1/32 and 1/16 of the trainset (including 165, 331 and 662 training images, respectively) as labeled data, and the remaining training images as unlabeled data for semi-supervised learning. The \textit{Baseline} and \textit{Oracle} are trained with supervised loss by using the split labelled training data and the whole trainset, respectively.}
\label{tab:voc}
\end{table}

\renewcommand\arraystretch{1.16}
\begin{table}[!ht]
\centering
\begin{footnotesize}
\begin{tabular}{p{2.0cm}|C{1.2cm}|*{3}{C{0.8cm}}}
\toprule
Method & Publication  &1/32  &1/16 &1/8  \\
\midrule
Baseline & -  &59.8  &64.3 &68.9  \\
GCT~\cite{ke2020guided} & ECCV 20  &-  &65.8 &71.3  \\
CCT~\cite{ouali2020semi}  & CVPR 20  &-  &66.4 &72.5 \\
DARS~\cite{he2021re} & ICCV 21  &61.9  &66.9 &73.7  \\
DBSN~\cite{Yuan_2021_ICCV} & ICCV 21   &62.2  &67.3 &73.5  \\
CAC~\cite{lai2021semi} & CVPR 21  &62.2  &69.4 &74.0  \\
CPS~\cite{chen2021semi} & CVPR 21  &62.5  &69.8 &74.4  \\
\textbf{USRN (Ours)} & -  &\textbf{64.6} &\textbf{71.2} &\textbf{75.0}  \\
\hline Oracle & -  &78.3  &78.3  &78.3  \\
\bottomrule
\end{tabular}
\end{footnotesize}
\caption{Quantitative comparison of USRN with the state-of-the-art over the dataset Cityscapes~\cite{cordts2016cityscapes}. We randomly split 1/32, 1/16 and 1/8 of the trainset (including 93, 186 and 372 training images, respectively) as labeled data, and the remaining training images as unlabeled data for semi-supervised learning. The \textit{Baseline} and \textit{Oracle} are trained with supervised loss by using the split labelled training data and the whole trainset, respectively.}
\label{tab:city}
\end{table}

\begin{figure*}[!t]
\centering
\begin{minipage}[h]{0.16\linewidth}
\centering\small {Input Image}
\end{minipage}
\begin{minipage}[h]{0.16\linewidth}
\centering\small {Ground Truth}
\end{minipage}
\begin{minipage}[h]{0.16\linewidth}
\centering\small {Baseline}
\end{minipage}
\begin{minipage}[h]{0.16\linewidth}
\centering\small {CAC~\cite{lai2021semi}}
\end{minipage}
\begin{minipage}[h]{0.16\linewidth}
\centering\small {CPS~\cite{chen2021semi}}
\end{minipage}
\begin{minipage}[h]{0.16\linewidth}
\centering\small {\textbf{USRN(Ours)}}
\end{minipage}
\centering
\begin{minipage}[h]{0.16\linewidth}
\centering\includegraphics[width=.99\linewidth]{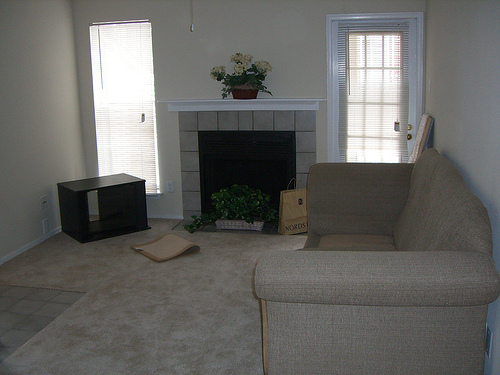}
\end{minipage}
\vspace{2pt}
\begin{minipage}[h]{0.16\linewidth}
\centering\includegraphics[width=.99\linewidth]{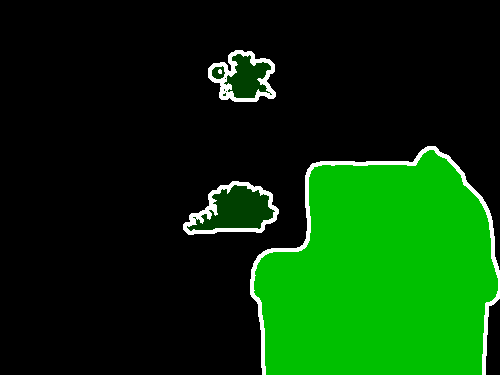}
\end{minipage}
\begin{minipage}[h]{0.16\linewidth}
\centering\includegraphics[width=.99\linewidth]{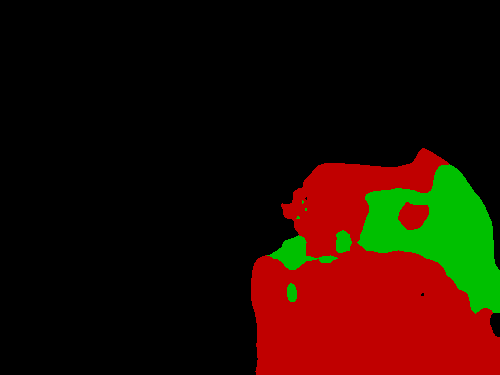}
\end{minipage}
\begin{minipage}[h]{0.16\linewidth}
\centering\includegraphics[width=.99\linewidth]{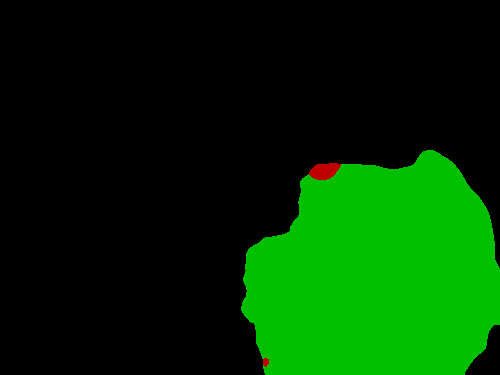}
\end{minipage}
\begin{minipage}[h]{0.16\linewidth}
\centering\includegraphics[width=.99\linewidth]{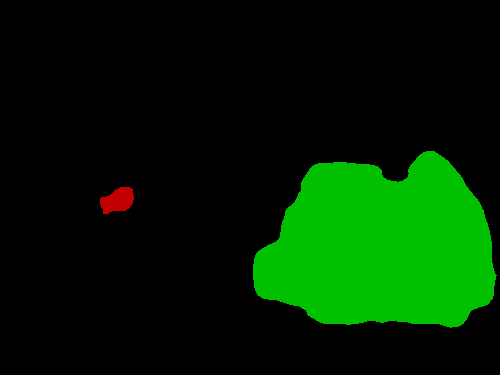}
\end{minipage}
\begin{minipage}[h]{0.16\linewidth}
\centering\includegraphics[width=.99\linewidth]{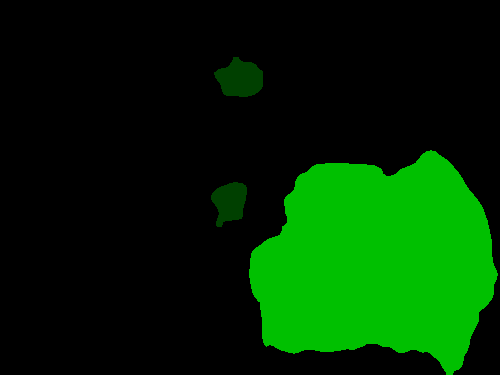}
\end{minipage}
\vspace{2pt}
\centering
\begin{minipage}[h]{0.16\linewidth}
\centering\includegraphics[width=.99\linewidth]{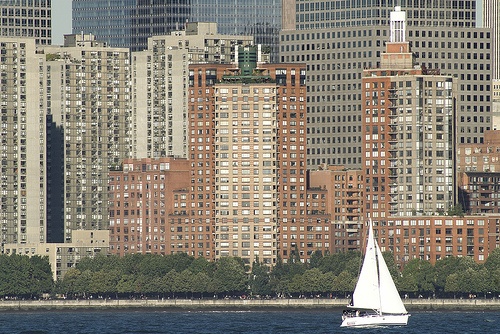}
\end{minipage}
\begin{minipage}[h]{0.16\linewidth}
\centering\includegraphics[width=.99\linewidth]{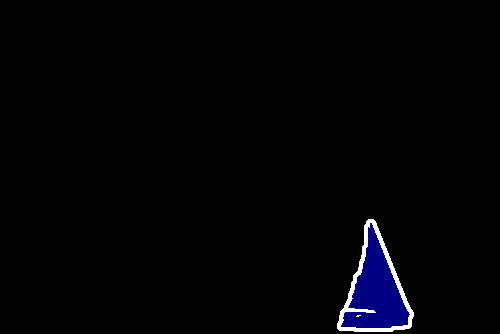}
\end{minipage}
\begin{minipage}[h]{0.16\linewidth}
\centering\includegraphics[width=.99\linewidth]{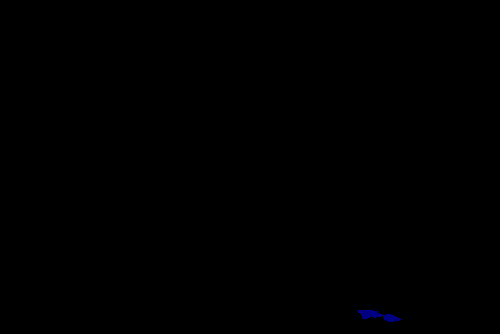}
\end{minipage}
\begin{minipage}[h]{0.16\linewidth}
\centering\includegraphics[width=.99\linewidth]{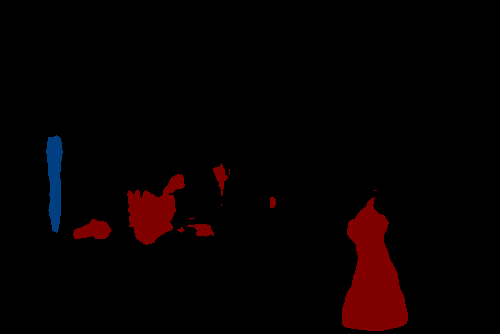}
\end{minipage}
\begin{minipage}[h]{0.16\linewidth}
\centering\includegraphics[width=.99\linewidth]{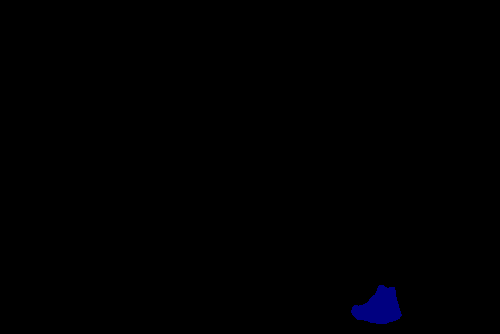}
\end{minipage}
\begin{minipage}[h]{0.16\linewidth}
\centering\includegraphics[width=.99\linewidth]{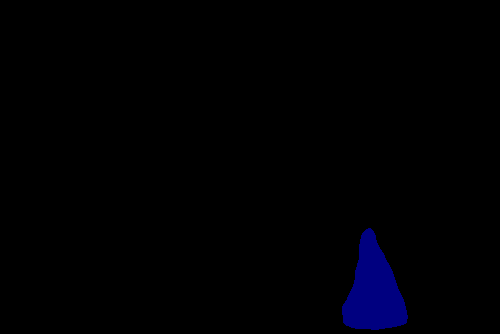}
\end{minipage}
\vspace{2pt}
\centering
\begin{minipage}[h]{0.16\linewidth}
\centering\includegraphics[width=.99\linewidth]{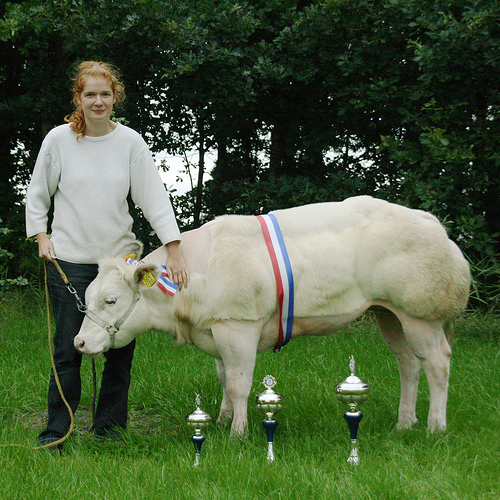}
\end{minipage}
\begin{minipage}[h]{0.16\linewidth}
\centering\includegraphics[width=.99\linewidth]{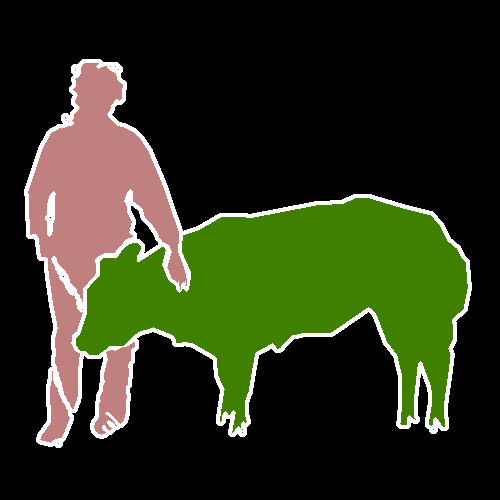}
\end{minipage}
\begin{minipage}[h]{0.16\linewidth}
\centering\includegraphics[width=.99\linewidth]{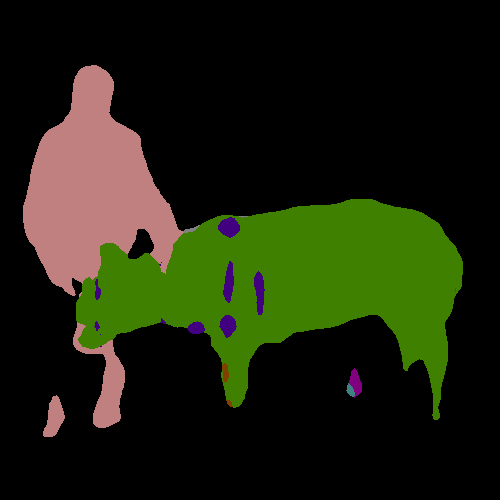}
\end{minipage}
\begin{minipage}[h]{0.16\linewidth}
\centering\includegraphics[width=.99\linewidth]{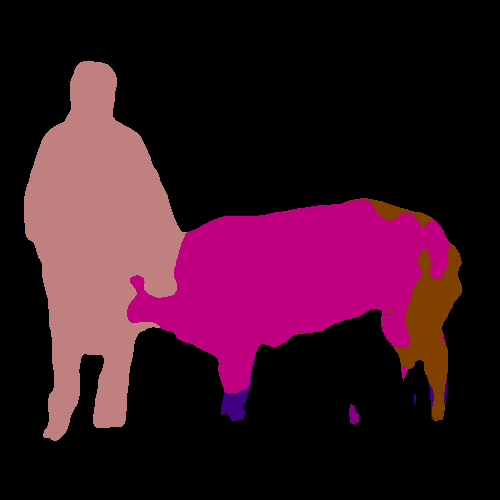}
\end{minipage}
\begin{minipage}[h]{0.16\linewidth}
\centering\includegraphics[width=.99\linewidth]{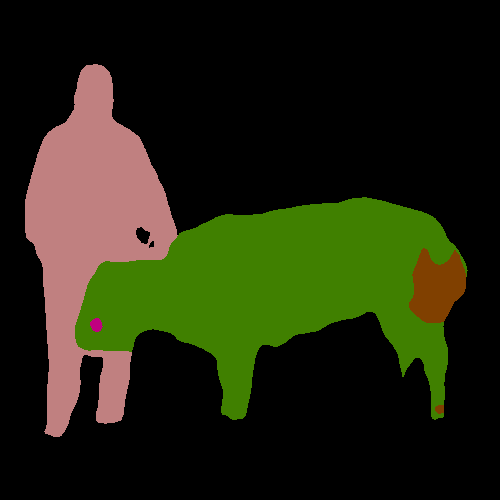}
\end{minipage}
\begin{minipage}[h]{0.16\linewidth}
\centering\includegraphics[width=.99\linewidth]{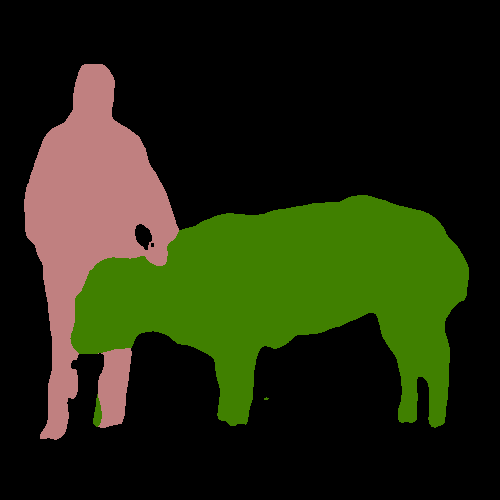}
\end{minipage}
\caption{Qualitative comparison of USRN with the state-of-the-art over 1/32 split of PASCAL VOC dataset. USRN can obtain more accurate semantic segmentation especially for pixels that are inaccurately segmented as the most dominant class (e.g., back-ground class as visualized in black color) by state-of-the-art methods~\cite{lai2021semi,chen2021semi}. }
\label{fig:comp}
\end{figure*} 

\subsection{Comparison with the state-of-the-art}
We compare USRN with state-of-the-art methods~\cite{ke2020guided,ouali2020semi, he2021re, Yuan_2021_ICCV,lai2021semi,chen2021semi} over PASCAL VOC and Cityscapes datasets~\cite{everingham2010pascal,cordts2016cityscapes}. Tables~\ref{tab:voc}~and~\ref{tab:city} shows experimental results. For PASCAL VOC dataset, we randomly split 1/64, 1/32 and 1/16 of the trainset (including 165, 331 and 662 training images, respectively) as labelled data, and the remaining training images as unllabeled data. As the number of training images in Cityscapes dataset is less than that in PASCAL VOC, we randomly split 1/32, 1/16 and 1/8 of the trainset (including 93, 186 and 372 training images, respectively) in Cityscapes dataset as labelled data, and the remaining training images as unlabelled data. State-of-the-art methods are implemented with various segmentation backbones and choose different splits of the trainset in experiments. For fair comparisons, we reproduce some experimental result by using the official code so that all the methods can be compared with the same split of labelled data as well as the same segmentation backbone.

As Tables~\ref{tab:voc}~and~\ref{tab:city} show, the proposed USRN outperforms the state-of-the-art consistently over the two datasets with different splits of labelled training data. The superior performance is largely attributed to the proposed unbiased subclass regularization that effectively addresses the class imbalance issue in semi-supervised segmentation. For smaller splits of the labelled training data, USRN outperforms the state-of-the-art with larger margins by 3.8\% and 2.1\% in mIoU for 1/64 split of PASCAL VOC and 1/32 split of Cityscapes, respectively. In particular, the performance of state-of-the-art methods is largely constrained by the quality of the segmentation model that is trained by using the class-imbalanced labelled data. Since deep convolutional neural networks tends to overfit with small datasets as proved in~\cite{srivastava2014dropout}, the class imbalance issue is more severe when training with fewer labelled data which degrades the performance of state-of-the-art methods greatly. While using larger splits of labelled data, the gaps between our method and the \textit{Oracle} as trained by using the whole trainset are 4.5\% in mIoU for 1/16 split of PASCAL VOC and 3.3\% in mIoU for 1/8 split of Cityscapes. Such experimental results show that our method can learn accurate segmentation models with a small amount of labelled training data, demonstrating its potential in reducing labelling efforts in deep network training.

We also provide qualitative comparisons over 1/32 split of PASCAL VOC dataset. We compare USRN with state-of-the-art methods~\cite{lai2021semi,chen2021semi} and the \textit{Baseline} that is trained with supervised loss only. The qualitative results are well aligned with the quantitative results as illustrated in Fig.~\ref{fig:comp}. It can be observed that USRN produces more accurate segmentations than state-of-the-art methods especially for those inaccurately segmented pixels that belong to the most dominant class. The qualitative experimental results further validate that USRN can better handle the class imbalance issue in semi-supervised semantic segmentation.

\subsection{Ablation studies}
We conducted extensive ablation studies to examine how the proposed USRN achieves the superior semi-supervised semantic segmentation. We performed all the ablation studies over 1/32 split of PASCAL VOC dataset, where USRN can achieve a mIoU of 68.6\% under default settings. Specifically, we examine different designs in USRN including different USRN components, different clustering strategies for class-balanced clusters generation, sharing features in different level (between the segmentation backbone $F$ and the auxiliary segmentation model $F_{sub}$), and parameter analysis of the confidence threshold $\gamma$ for pseudo label selection. 

\renewcommand\arraystretch{1.16}
\begin{table}[!ht]
\centering
\begin{footnotesize}
\begin{tabular}{C{1.5cm}|*{5}{C{0.6cm}}|C{0.6cm}}
\toprule
Model &MSL &OST &USR &SST &EGM &mIoU \\
\midrule
Model~\RN{1} &\checkmark & & &  &  &59.1 \\
Model~\RN{2} &\checkmark &\checkmark & &  &   &64.1 \\
Model~\RN{3} &\checkmark &\checkmark &\checkmark &  &   &65.0 \\
Model~\RN{4} &\checkmark &\checkmark &\checkmark &\checkmark  &   &67.1 \\
\textbf{USRN} &\checkmark &\checkmark &\checkmark &\checkmark  &\checkmark  &68.6 \\
\bottomrule
\end{tabular}
\end{footnotesize}
\caption{Ablation study on different components (\ie, MSL, OST, USR, SST and EGM) of USRN over 1/32 split of PASCAL VOC dataset. Here, MSL, OST, USR, SST and EGM are abbreviations of multi-distribution supervised learning, original self-training, unbiased subclass regularization, subclass self-training and entropy-base gate mechanism, respectively.}
\label{tab:abla}
\end{table}

\noindent\textbf{Different Components.} We conducted ablation studies on different components of USRN to examine their effectiveness as shown in Table.~\ref{tab:abla}. Specifically, we trained five models over 1/32 split of PASCAL VOC dataset including: 1) \textbf{Model~\RN{1}} that is trained with labeled data only using the multi-distribution supervised learning (MDL) loss $\mathcal{L}_{s}^{md}$ in Eq.~\ref{loss:md}; 2) \textbf{Model~\RN{2}} that performs self-training on original class distributions only using the MDL loss and the original self-training (OST) loss as in~\cite{sohn2020fixmatch,lai2021semi,chen2021semi}; 3) \textbf{Model~\RN{3}} that performs unbiased subclass regularization (USR) directly on the OST in \textbf{Model~\RN{2}} by using the MDL loss and the proposed self-training loss $\mathcal{L}_{st}$ as defined in Eq.~\ref{loss:st}; 4) \textbf{Model~\RN{4}} that includes subclass self-training (SST) loss in Eq.~\ref{loss:sub_st} into \textbf{Model~\RN{3}} for training the auxiliary segmentation model on unlabelled data; and 5) \textbf{USRN} that introduces the entropy-based gate mechanism (EGM) into \textbf{Model~\RN{4}} to coordinate the concurrent learning from the original classes and the generated subclasses.

As Table~\ref{tab:abla} shows, both \textbf{Model~\RN{2}} and \textbf{Model~\RN{3}} outperform \textbf{Model~\RN{1}} by large margins, demonstrating the effectiveness of self-training in semi-supervised segmentation. Without SST, the performance of \textbf{Model~\RN{3}} still outperforms \textbf{Model~\RN{2}}, which shows that the subclass segmentation model trained with the labelled data only can produce high-quality pseudo labels on unlabelled data. With SST, \textbf{Model~\RN{4}} outperforms \textbf{Model~\RN{3}} by 2.1\% in mIoU thanks to updating the subclass segmentation model by self-training on unlabelled data. With the updated auxiliary segmentation model, more accurate subclass segmentation can be produced to generate unbiased pseudo labels for updating the segmentation backbone. Finally, \textbf{USRN} further improves \textbf{Model~\RN{4}} by 1.5\% in mIoU, which validates the effectiveness of the proposed entropy-based gate mechanism.

\renewcommand\arraystretch{1.16}
\begin{table}[!ht]
\centering
\begin{footnotesize}
\begin{tabular}{C{2.5cm}|C{1.6cm}C{1.8cm}C{0.6cm}}
\toprule
{Clustering algorithm} &{Original CBR}  &{Subclass CBR}  &{mIoU} \\
\midrule
{Normal k-means} &{33.8\%} &{96.4\%} &{68.0} \\
{Balanced k-means} &{33.8\%} &{\textbf{99.5\%}} &{\textbf{68.6}} \\
\bottomrule
\end{tabular}
\end{footnotesize}
\caption{Comparisons of normal k-means~\cite{macqueen1967some} with balanced k-means~\cite{lin2019balanced} while applying USRN to 1/32 split of PASCAL VOC dataset. Here, the class balance rate (CBR) is defined in Eq.~\ref{eq:cb}, where CBR=100\% means the number of pixels within each class is equal (\ie, extreme class balance) and CBR=0\% means all pixels are labeled with only one class (\ie, extreme class imbalance).}
\label{tab:kmeans}
\end{table}

\noindent\textbf{Clustering Strategy.}
In Section.~3.2, we adopted balanced k-means clustering~\cite{lin2019balanced} to generate class-balanced subclass annotations. To measure the class balance of annotations, we define a new metric named class balance rate (CBR) which can be formulated as follows:
\begin{equation}
\text{CBR} = 1- \dfrac{\sigma_{c}}{\sigma_{c}^\star} =  1- \dfrac{\sqrt{\dfrac{1}{C}\sum\limits_{n=1}^{C}n_{c}^{2}-(\dfrac{1}{C}\sum\limits_{n=1}^{C}n_{c})^{2}}}
{\sqrt{\dfrac{1}{C}(\sum\limits_{n=1}^{C}n_{c})^{2}-(\dfrac{1}{C}\sum\limits_{n=1}^{C}n_{c})^{2}}},
\label{eq:cb}
\end{equation}
where $n_{c}$ is the number of pixels within each class $c \subset (1, C)$ for given annotations, 
$\sigma_{c}$ is standard deviation of $\{n_{1}, n_{2}, \cdots, n_{C} \}$ and $\sigma_{c}^\star$ is standard deviation of $\{\sum\limits_{n=1}^{C}n_{c}, 0, \cdots, 0 \}$, \ie, all pixels are labeled with only one class (extreme class imbalance).

As shown in Table~\ref{tab:kmeans}, the CBR of subclass annotations is almost 100\% which is much higher than the CBR of the original annotations. This demonstrates that we obtain class-balanced subclass annotations from class-imbalanced original annotations successfully. We can also observe that the subclass annotations generated with normal k-means~\cite{macqueen1967some} is also quite class-balanced (CBR=96.4\%), and USRN model trained with such annotations can achieve comparable accuracy as the USRN trained with default clustering strategy (\ie, balanced k-means). This shows that our method is robust to different clustering strategies.

\renewcommand\arraystretch{1.16}
\begin{table}[!ht]
\centering
\begin{footnotesize}
\begin{tabular}{C{4.3cm}|C{2.0cm}C{0.6cm}}
\toprule
{Sharing Features} &{GPU Occupation} &{mIoU} \\
\midrule
No sharing &9.76 Gb$\times{}$2 &67.3 \\
Low-level features &8.75 Gb$\times{}$2 &\textbf{68.6} \\
Both low-level and high-level features &6.99 Gb$\times{}$2 &67.8 \\
\bottomrule
\end{tabular}
\end{footnotesize}
\caption{The impact of feature sharing between the segmentation backbone $F$ and the auxiliary network $F_{sub}$ over 1/32 split of PASCAL VOC dataset: USRN trained with default setting (\ie, sharing low-level features) achieved the best mIoU with a little computation overhead during training. 
Note that the computational cost is equal for all the settings during inference.
}
\label{tab:share}
\end{table}

\noindent\textbf{Sharing Features.} Recent supervised segmentation models~\cite{zhao2017pyramid,badrinarayanan2017segnet,zhao2018psanet,chen2018encoder} achieved high accuracy by integrating multi-level features. In default setting of USRN, the segmentation backbone $F$ and the auxiliary segmentation model $F_{sub}$ share layers that extract low-level features. We further evaluate the impact of sharing features between $F$ and $F_{sub}$. As shown in Table~\ref{tab:share}, USRN trained with default setting (\ie, sharing low-level features) achieves the highest accuracy in mIoU as compared with USRN trained with other settings (\ie, `no sharing' and sharing multi-level features). The setting of `no sharing' has the lowest accuracy, which demonstrates that original class segmentation and auxiliary subclass segmentation are complementary to each other. The reason why sharing high-level features (\ie, semantic features) degrades the accuracy of USRN is that original class segmentation and auxiliary subclass segmentation require to learn different semantic features as semantic information of these two tasks is different.

\renewcommand\arraystretch{1.16}
\begin{table}[!ht]
\centering
\begin{footnotesize}
\begin{tabular}{C{1.0cm}|*{6}{C{0.7cm}}}
\toprule
$\gamma$ &0.55 &0.65 &0.75 &0.85 &0.95 &0.99 \\
\midrule
mIoU &67.4 &67.7 &68.6 &68.6 &68.5 &68.1 \\
\bottomrule
\end{tabular}
\end{footnotesize}
\caption{Sensitivity of the confidence threshold $\gamma$ in Eq.~\ref{eq:s}: USRN is stable when $\gamma$ changes in a range from 0.75 to 0.95. The experiments are conducted over 1/32 split of PASCAL VOC dataset.}
\label{tab:para}
\end{table}

\renewcommand\arraystretch{1.16}
\begin{table*}[!ht]
\centering
\begin{footnotesize}
\begin{tabular}{p{1.6cm}|*{21}{C{0.28cm}}p{0.5cm}}
 \toprule
 Method  &back. &aero. &bicy. &bird &boat &{bott.} &bus &car &cat &chair &{cow} &table &dog &horse &motor &pers. &plant &{sheep} &sofa &train &tv &mIoU \\
 \midrule
 Baseline &89.9 &73.6 &33.8 &75.1 &42.0 &54.4 &80.0 &75.8 &78.9 &24.7 &50.2 &43.1 &72.6 &50.2 &68.2 &77.2 &34.9 &64.8 &30.6 &67.6 &55.1 &59.2  \\
 CReST~\cite{wei2021crest} &90.5 &77.0 &\textbf{38.6} &74.8 &48.2 &52.1 &83.3 &76.0 &82.9 &24.9 &61.2 &49.8 &79.6 &63.7 &71.2 &77.3 &41.5 &65.9 &34.8 &74.7 &59.1 &63.2  \\
 DARS~\cite{he2021re} &91.3 &82.6 &37.4 &81.9 &50.5 &58.6 &\textbf{88.5} &\textbf{82.9} &82.8 &25.5 &56.3 &49.1 &75.3 &64.6 &73.6 &79.7 &\textbf{42.2} &64.0 &37.1 &73.4 &57.9 &64.5  \\
 USRN (Ours) &\textbf{91.9} &\textbf{84.1} &36.1 &\textbf{84.9} &\textbf{52.8} &\textbf{66.4} &87.9 &81.8 &\textbf{86.4} &\textbf{26.5} &\textbf{75.2} &\textbf{58.6} &\textbf{83.0} &\textbf{73.3} &\textbf{74.7} &\textbf{80.2} &40.7 &\textbf{76.2} &\textbf{42.0} &\textbf{78.5} &\textbf{59.8} &\textbf{68.6} \\
\bottomrule
\end{tabular}
\end{footnotesize}
\vspace{-4pt}
\caption{Quantitative comparisons of USRN with multiple class-imbalance learning methods for semi-supervised semantic segmentation. The experiments are conducted over 1/32 split of PASCAL VOC dataset.}
\label{tab:cb}
\end{table*}

\noindent\textbf{Parameter Analysis.} The confidence threshold $\gamma$ in Eq.~\ref{eq:s} is an important hyper-parameter for generating high-quality class-unbiased pseudo labels. We evaluate USRN with different $\gamma$ and Table~\ref{tab:para} show experimental results. It can be observed that USRN is very stable when $\gamma$ changes in a range from 0.75 to 0.95. While $\gamma$ is smaller than 0.75, the performance of USRN degrades because the predicted pseudo labels tend to become noisy. While $\gamma$ is larger than 0.95, USRN suffers from over-fitting because the very high confidence threshold returns very limited pseudo labels. We set $\gamma$ at 0.75 by default in our implemented USRN.

\subsection{Discussion}
\noindent\textbf{Comparison with Class-Imbalance Methods:} The proposed USRN explores class-unbiased segmentation to address the class imbalance issue in semi-supervised segmentation. Recently, several studies~\cite{wei2021crest,he2021re} attempt to handle the class imbalance issue in semi-supervised learning. We compare USRN with these methods and Table~\ref{tab:cb} shows experimental results. It can been seen that USRN achieves the best overall performance (\ie, 68.6 in mIoU) and the best per-class accuracy on 17 out of all 21 classes. The superior performance shows that exploring class-unbiased segmentation from balanced subclass distributions is more effective than selecting more pseudo labels for minority classes in self-training as in~\cite{wei2021crest,he2021re}.

\renewcommand\arraystretch{1.16}
\begin{table}[!ht]
\centering
\begin{footnotesize}
\begin{tabular}{C{2cm}|*{3}{C{1.0cm}}}
\toprule
{Method} &{Base}  &{+ USRN}
&{Gain} \\
\midrule
DARS~\cite{he2021re} &64.5 &69.0 &+4.5 \\
CPS~\cite{chen2021semi}  &64.8 &69.2 &+4.4 \\
CAC~\cite{lai2021semi}  &65.1 &70.0 &+4.9 \\
\bottomrule
\end{tabular}
\end{footnotesize}
\caption{The proposed USRN complements with state-of-the-art methods~\cite{he2021re,chen2021semi,lai2021semi} 
over 1/32 split of PASCAL VOC dataset: the performance of all tested state-of-the-art methods can be improved greatly with the integration of USRN. }
\label{tab:comp}
\end{table}

\noindent\textbf{Complementary Studies:} We also investigate whether the proposed USRN can complement with state-of-the-art methods~\cite{he2021re,lai2021semi,chen2021semi} as compared in Section~4.2. We integrate our proposed unbiased subclass regularization networks into the state-of-the-art methods to perform this study. Table \ref{tab:comp} shows experimental results. It can be observed that the integration of USRN improves performance greatly across all tested state-of-the-art methods which employ either consistency-training~\cite{lai2021semi} or self-training~\cite{he2021re,chen2021semi}.

\renewcommand\arraystretch{1.16}
\begin{table}[!ht]
\centering
\begin{footnotesize}
\begin{tabular}{C{2cm}|*{3}{C{1.0cm}}}
\toprule
{Architecture} &{Baseline}  &{USRN}
&{Gain} \\
\midrule
PSPNet~\cite{zhao2017pyramid} &49.7 &65.4 &+15.7 \\
PSANet~\cite{zhao2018psanet} &56.5 &66.5 &+10.0 \\
Deeplabv3+~\cite{chen2018encoder} &59.2 &68.6 &+9.4 \\
\bottomrule
\end{tabular}
\end{footnotesize}
\caption{The proposed USRN can work well with different semantic segmentation architectures~\cite{zhao2017pyramid,zhao2018psanet,chen2018encoder} with significant performance improvement as compared with the \textit{Baseline}
over 1/32 split of PASCAL VOC dataset.
}
\label{tab:arch}
\end{table}

\noindent\textbf{Different Segmentation Architectures:} We further study whether USRN can work well with different semantic segmentation architectures. We studied three widely used segmentation architectures including PSPNet~\cite{zhao2017pyramid}, PSANet~\cite{zhao2018psanet} and Deeplabv3+~\cite{chen2018encoder} and Table \ref{tab:arch} shows experimental results. It can be observed that the proposed USRN outperforms the \textit{Baseline} model with large margins consistently with the three architectures. This shows that USRN can work well with different semantic segmentation architectures that apply pyramid spatial pooling~\cite{zhao2017pyramid}, attention mechanism~\cite{zhao2018psanet} and dilated convolutions~\cite{chen2018encoder}.

\section{Conclusion}
This paper presents an unbiased subclass regularization network that explores class-unbiased segmentation to address class imbalance issue in semi-supervised segmentation. Specifically, the class-biased segmentation learnt in imbalanced original class distributions is regularized by the class-unbiased segmentation learnt in balanced subclass distributions. To coordinate the concurrent learning from the original class and the generated subclass, an entropy-based gate mechanism is designed to suppress unconfident subclass predictions for facilitating subclass regularization. Comprehensive experiments demonstrate the effectiveness of our method in semi-supervised segmentation. In the future, we will investigate how the idea of unbiased subclass regularization perform in other semi-supervised learning tasks such as semi-supervised image classification and semi-supervised object detection.

\textbf{Acknowledgement.} 
This study is supported under the RIE2020 Industry Alignment Fund – Industry Collaboration Projects (IAF-ICP) Funding Initiative, as well as cash and in-kind contribution from Singapore Telecommunications Limited (Singtel), through Singtel Cognitive and Artificial Intelligence Lab for Enterprises (SCALE@NTU).

{\small
\bibliographystyle{ieee_fullname}
\bibliography{manuscript}
}

\end{document}